\documentclass[times,twocolumn,final]{elsarticle}

\usepackage{framed,multirow}

\usepackage{amssymb}
\usepackage{latexsym}

\usepackage{multirow}
\usepackage{graphicx}
\usepackage{booktabs}
\usepackage{caption}
\usepackage{array}
\usepackage{makecell}

\usepackage{url}
\usepackage{xcolor}

\usepackage{pifont}

\usepackage[hidelinks]{hyperref}
\definecolor{newcolor}{rgb}{.8,.349,.1}

\usepackage[ruled,vlined,algo2e]{algorithm2e}
\SetKwComment{Comment}{$\triangleright$\ }{}

\usepackage{mathtools}
\usepackage{amsmath}

\begin{document}


\begin{frontmatter}

\title{Sparser2Sparse: Single-shot Sparser-to-Sparse Learning for Spatial Transcriptomics Imputation with Natural Image Co-learning}

\author[1]{Yaoyu Fang}
\ead{yaoyu.fang@northwestern.edu}
\author[1]{Jiahe Qian}
\author[2]{Xinkun Wang}
\author[3]{Lee A. Cooper}
\author[1]{Bo Zhou}
\ead{bo.zhou@northwestern.edu}

\address[1]{Department of Radiology, Northwestern University, Chicago, IL, USA}
\address[2]{Department of Cell and Developmental Biology, Northwestern University, Chicago, IL, USA}
\address[3]{Department of Pathology, Northwestern University, Chicago, IL, USA}

\begin{abstract}
Spatial transcriptomics (ST) has revolutionized biomedical research by enabling high resolution gene expression profiling within tissues. However, the high cost and scarcity of high resolution ST data remain significant challenges. We present Single-shot Sparser-to-Sparse (S2S-ST), a novel framework for accurate ST imputation that requires only a single and low-cost sparsely sampled ST dataset alongside widely available natural images for co-training. Our approach integrates three key innovations: (1) a sparser-to-sparse self-supervised learning strategy that leverages intrinsic spatial patterns in ST data, (2) cross-domain co-learning with natural images to enhance feature representation, and (3) a Cascaded Data Consistent Imputation Network (CDCIN) that iteratively refines predictions while preserving sampled gene data fidelity. Extensive experiments on diverse tissue types, including breast cancer, liver, and lymphoid tissue, demonstrate that our method outperforms state-of-the-art approaches in imputation accuracy. By enabling robust ST reconstruction from sparse inputs, our framework significantly reduces reliance on costly high resolution data, facilitating potential broader adoption in biomedical research and clinical applications.

\end{abstract}

\begin{keyword}
Spatial Transcriptomics, Gene Expression Imputation, Single-shot Learning, Natural Image Co-training, Cost Reduction
\end{keyword}

\end{frontmatter}


\section{Introduction}
Spatial transcriptomics (ST) is a cutting-edge technology that enables the investigation of spatially resolved gene expression within tissues \citep{aspSpatiallyResolvedTranscriptomes2020}. Traditional transcriptomic approaches, such as single-cell RNA sequencing (scRNA-seq), provide high-throughput, high resolution gene expression profiles but inherently lack spatial context \citep{aungSpatiallyInformedGene2024,boeSpatialTranscriptomicsReveals2024,sankar50480ImmuneInfiltrate2024}. However, spatial information is crucial for identifying disease biomarkers, understanding disease progression, and developing personalized treatment strategies.

ST has emerged as a transformative tool across multiple biomedical fields, offering unprecedented insights into tissue organization and function. In oncology, it enhances our understanding of tumor heterogeneity and immune responses, facilitating cancer diagnosis and treatment by identifying novel cell types and immune correlates \citep{wangSpatialTranscriptomicsProteomics2021,liSpatialTranscriptomicsTumor2022}. For instance, in breast cancer, ST has uncovered distinct gene expression patterns across tumor regions, enabling precise classification of subtypes and guiding targeted therapies \citep{levy-jurgensonSpatialTranscriptomicsInferred2020,coutantSpatialTranscriptomicsReveal2023,anSpatialTranscriptomicsBreast2024}. In melanoma, it provides critical spatial insights into immune cell infiltration, essential for predicting and optimizing immunotherapy responses \citep{boeSpatialTranscriptomicsReveals2024,sankar50480ImmuneInfiltrate2024,aungSpatiallyInformedGene2024}. In neurology, ST advances brain mapping and neurodegenerative disease research, aiding in neural classification and biomarker discovery for conditions like Alzheimer’s and Parkinson’s \citep{chenSpatialTranscriptomicsSitu2020,navarroSpatialTranscriptomicsReveals2020,jungSpatialTranscriptomicsNeuroscience2023,piweckaSinglecellSpatialTranscriptomics2023,heUnravelingAlzheimersDisease2024,zhangInvestigatingMechanismsInflammation2024}. It has revealed spatial patterns of neuroinflammation and protein aggregation in Alzheimer’s disease \citep{chenSpatialTranscriptomicsSitu2020,navarroSpatialTranscriptomicsReveals2020,heUnravelingAlzheimersDisease2024} and identified distinct molecular signatures in the substantia nigra in Parkinson’s disease, potentially enabling earlier interventions \citep{zhangInvestigatingMechanismsInflammation2024}. In cardiology, ST redefines our understanding of heart repair mechanisms and coronary atherosclerosis, supporting precision medicine approaches \citep{rothSinglecellSpatialTranscriptomics2020,boileauFullLengthSpatialTranscriptomics2022,kuppeSpatialMultiomicMap2022,longSinglecellSpatialTranscriptomics2023}. Studies have uncovered spatial gene expression patterns in pathological remodeling and identified distinct zones of repair and regeneration in infarcted heart tissue, leading to novel therapeutic targets for cardiac repair \citep{boileauFullLengthSpatialTranscriptomics2022,kuppeSpatialMultiomicMap2022}. Beyond these fields, ST has broad applications in reproductive biology, immunology, and developmental biology, providing insights into tissue patterning and immune interactions \citep{anderssonSpatialDeconvolutionHER2positive2021}. Its integration with multiomics data further amplifies its potential for future research and clinical applications \citep{driesAdvancesSpatialTranscriptomic2021,kleinoComputationalSolutionsSpatial2022,duAdvancesSpatialTranscriptomics2023}. By continually revealing new dimensions of tissue organization and function, ST challenges existing paradigms and unlocks new therapeutic opportunities, solidifying its role as a cornerstone of modern biomedical research.

Despite its transformative potential across diverse biomedical fields, ST faces several critical challenges that hinder its broader adoption in clinical and research settings. The first major barrier is the substantial financial and logistical burden associated with dense tissue spot sampling\citep{fangComputationalApproachesChallenges2023,smithChallengesOpportunitiesClinical2024}. While advances in ST platforms, such as 10x Genomics Visium HD and Xenium, have significantly improved resolution, achieving over 50,000 spots per section at a 2-micron spot diameter, these improvements come at a steep cost. A single Xenium experiment typically costs $2,000 to $4,000, excluding additional expenses for labor, reagents, and data storage, while platforms such as Visium HD require additional costs for sequencing and library preparation. Such high costs pose a major constraint on large-scale studies involving hundreds of samples, limiting ST’s feasibility for clinical and translational research. Beyond financial constraints, the sheer volume of data generated by high resolution platforms presents another formidable challenge\citep{fangComputationalApproachesChallenges2023,smithChallengesOpportunitiesClinical2024}. Xenium experiments, for instance, can produce terabytes of data per sample, requiring substantial computational and storage infrastructure that many research laboratories and clinical facilities lack. This scalability issue further hinders ST’s widespread implementation. Moreover, due to these financial and technical barriers, large-scale ST datasets remain prohibitively expensive to generate. Compounding this issue, most existing ST datasets are proprietary, restricting open access and data sharing. This scarcity of publicly available, high quality data creates a significant bottleneck for AI-driven solutions, which typically rely on extensive training datasets. These limitations collectively underscore the urgent need for cost-effective approaches and technological innovations to enhance ST’s accessibility, scalability, and practical utility in biomedical research and clinical applications.

For low resolution ST systems like Visium (55-$\mu$m spot diameter, in contrast to 2-$\mu$m spots in Visium HD and 0.2-$\mu$m optical resolution in Xenium), numerous efforts have been devoted to improving spatial resolution. These approaches generally fall into two categories: histology-based and histology-free (ST-only) methods. Histology-based methods leverage ST-paired histological features to predict gene expression. For instance, DeepSpaCE \citep{monjoEfficientPredictionSpatial2022} introduced a convolutional neural network (CNN) to infer gene-expression profiles from H\&E-stained images, while EGN \citep{yangExemplarGuidedDeep2022} employed a vision transformer-based cascade to enhance long-range relationship modeling. Similarly, XFUSE \citep{bergenstrahleSuperresolvedSpatialTranscriptomics2022} developed a variational autoencoder (VAE)-based deep generative model for gene expression inference, and iStar \citep{zhangInferringSuperresolutionTissue2024} combined a pre-trained hierarchical histology feature extractor \citep{chenScalingVisionTransformers2022} with a fine-tuned multilayer perceptron (MLP) to improve resolution. In addition to deep learning, methods like TESLA \citep{huDecipheringTumorEcosystems2023} use neighborhood histological similarity for gene expression interpolation, though deep learning generally outperforms such traditional approaches. Histology-free methods, on the other hand, infer high resolution ST data without relying on histological inputs. BayesSpace \citep{zhaoSpatialTranscriptomicsSubspot2021} employs a Bayesian statistical framework to infer sub-spot gene expression based on spatial neighborhood values, while GNTD \citep{songGNTDReconstructingSpatial2023} constructs a spatial-transcriptomic graph that integrates spatial and gene expression data, which is then processed through an MLP to reconstruct high resolution ST. Our work aligns with this histology-independent category, which is advantageous because histological inputs can be inconsistent, subject to variations in staining protocols, imaging devices, and acquisition conditions.

Despite these advancements, key challenges remain. First, histology-based methods inherently depend on histological inputs, which may be difficult to obtain and highly variable. Second, existing approaches often treat ST resolution enhancement as a spot-by-spot prediction from histology, neglecting the broader spatial context and long-range relationships between ST spots—critical factors for effective imputation learning. Third, deep learning models require large-scale, diverse datasets, yet high-quality ST data is scarce, particularly for specific tissue types. Data-sharing restrictions further exacerbate this issue, making sample-specific, single-shot learning highly desirable. Fourth, most existing methods adopt standard computer vision architectures without fully considering the unique characteristics of ST data, leaving room for application-specific model designs. Most importantly, prior research has largely focused on enhancing resolution from low resolution ST data, while the crucial challenge of reducing the number of required spot samples to lower experimental costs, without sacrificing resolution, remains underexplored. Given the prohibitive costs of ultra-high-resolution ST systems like Xenium and Visium HD, addressing this gap is essential for making these technologies more cost-effective and scalable for broader biomedical applications.

To address the aforementioned challenges, we propose a novel single-shot sparser-to-sparse learning framework with natural image co-learning (S2S-ST) for cost-effective, high resolution ST imputation from sparse samples. S2S-ST introduces three key innovations: \textbf{(1) Sparser-to-Sparse Learning for Ultra-High-Resolution ST Imputation}: We consider a scenario where only a sparse subset of ultra-high-resolution ST data is available and aim to recover the full resolution gene expression profile. To achieve this, we introduce an innovative sparser-to-sparse framework, enabling single-shot imputation learning using only the sample-specific sparse data without requiring large external datasets. \textbf{(2) Natural Image Co-learning for Enhanced Representation}:
Given the challenge of limited data representation when only a single ST sample is available for imputation learning, we propose a natural image imputation co-learning strategy. By leveraging structural similarities between natural images and spatial transcriptomics data, this strategy enhances the model’s ability to learn spatial patterns, improving imputation performance. \textbf{(3) Cascade Data-Consistent Imputation Network (CDCIN)}: To ensure biologically consistent imputation, we design a customized cascade data-consistent imputation network (CDCIN). This architecture incorporates a powerful image restoration backbone combined with a data consistency layer, preserving the already acquired high resolution ST spots while optimizing the imputed values for missing regions. We validated S2S-ST on a diverse set of high resolution ST samples and demonstrated that it can generate high quality ST profiles from sparsely sampled data, significantly reducing the cost of ST experiments. Our method outperforms competitive baseline approaches, providing an efficient and accurate solution for sparse ST imputation. We believe that S2S-ST represents a major step toward making ultra-high-resolution ST more accessible and cost-effective, facilitating broader adoption in biomedical research and clinical applications.

\section{Methods}
\begin{figure*}[htb!]
    \centering
    \includegraphics[width=0.98\textwidth]{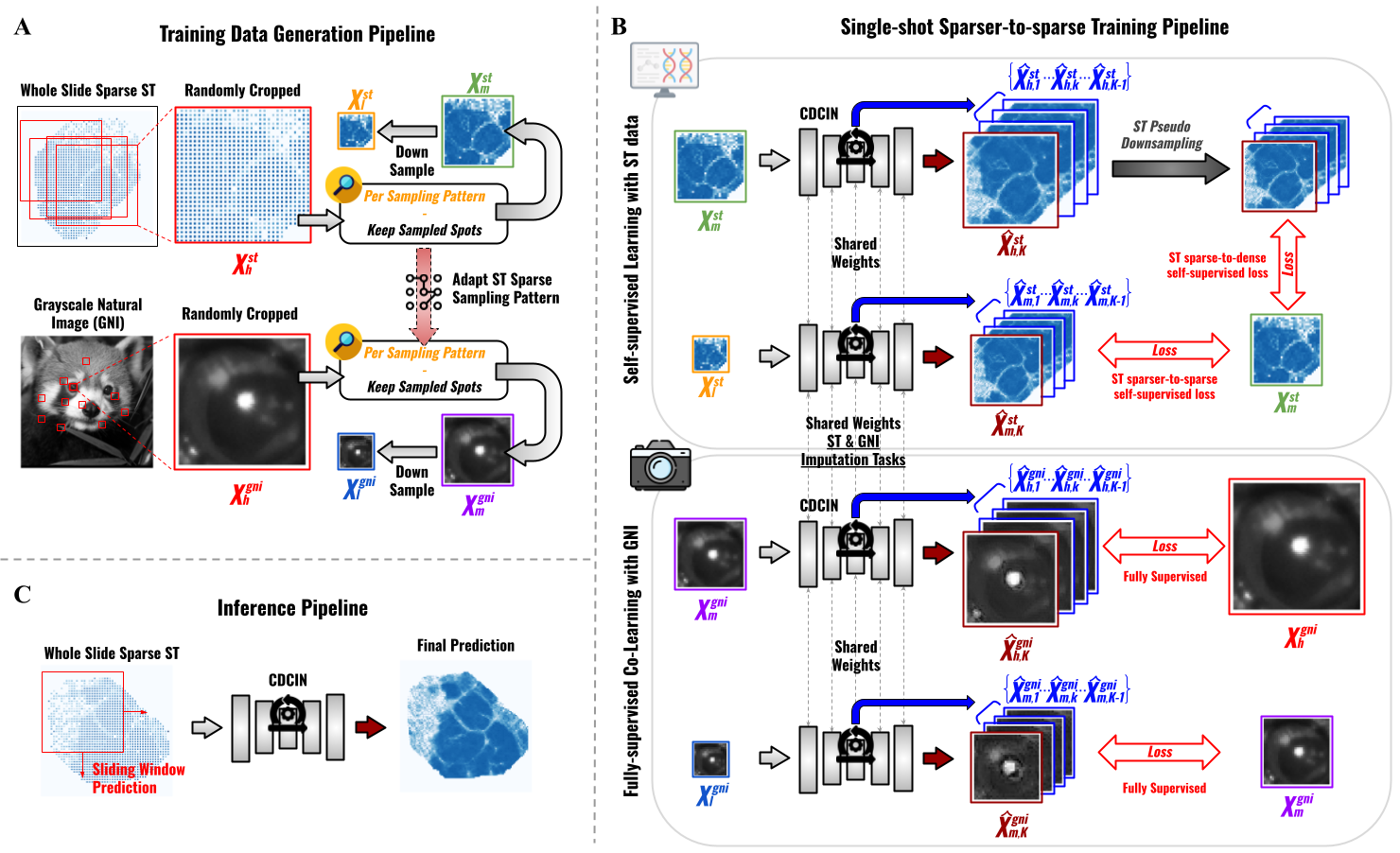}
    \caption{Overview of the Single-shot S2S-ST framework. 
    \textbf{(A) Training data generation}: A single whole slide sparse sampled ST dataset is randomly cropped into high resolution patches ($X_h^{st}$), which are downsampled using binary sampling masks to generate medium and low resolution inputs ($X_m^{st}, X_l^{st}$). Grayscale natural images (GNI) are processed in parallel using the same sampling patterns, producing matched inputs ($X_h^{gni}, X_m^{gni}, X_l^{gni}$) for cross-domain learning. 
    \textbf{(B) Single-shot sparser-to-sparse training}: A shared Cascaded Data Consistent Imputation Network (CDCIN) is trained jointly on ST and GNI data. For ST, two self-supervised losses are used: a sparser-to-sparse loss on $X_m^{st}$ and a sparse-to-dense downsampling loss on $X_h^{st}$. For GNI, fully-supervised learning is performed to strengthen feature representation. 
    \textbf{(C) Inference}: The trained CDCIN predicts full resolution expression maps from sparse ST inputs using a sliding window approach, with overlapping regions aggregated via weighted averaging. This framework enables accurate and biologically consistent gene expression recovery from limited ST data, enhanced by natural image co-learning.}
    \label{fig:model-arch}
\end{figure*}

The overall framework of our proposed method is illustrated in Fig.~\ref{fig:model-arch}. Our framework consists of three key components designed to address the challenges of overfitting and data scarcity in single-sample ST imputation tasks. First, we designed a \textbf{single-shot sparser-to-sparse self-supervised learning framework} (Section~\ref{subsec:image-co-train}), which jointly trains on a single ST sample in a self-supervised manner and large-scale natural images in a fully-supervised manner to enhance imputation capabilities. Second, we propose a \textbf{cascaded data consistent imputation network (CDCIN)} (Section~\ref{subsec:cascade}), a ST task-specific network that ensures high quality and high resolution ST data recovery through iterative refinement. Finally, within the CDCIN, we design a powerful gene restoration network, called \textbf{residual dense hybrid attention network (RDHAN)} (Section~\ref{subsec:RDHAN}), which integrates channel and spatial attention mechanisms to enhance gene expression imputation accuracy. Technical details and implementation of each component are described in the following sections.

We also provide the detailed implementation of our framework (Section~\ref{subsec:implement}), as well as the datasets (Section~\ref{subsec:datasets}), baseline methods , and evaluation metrics (Section~\ref{subsec:baseline+metrics}) used in our experiments.

\begin{figure*}[htb] 
    \centering
    \includegraphics[width=0.95\textwidth]{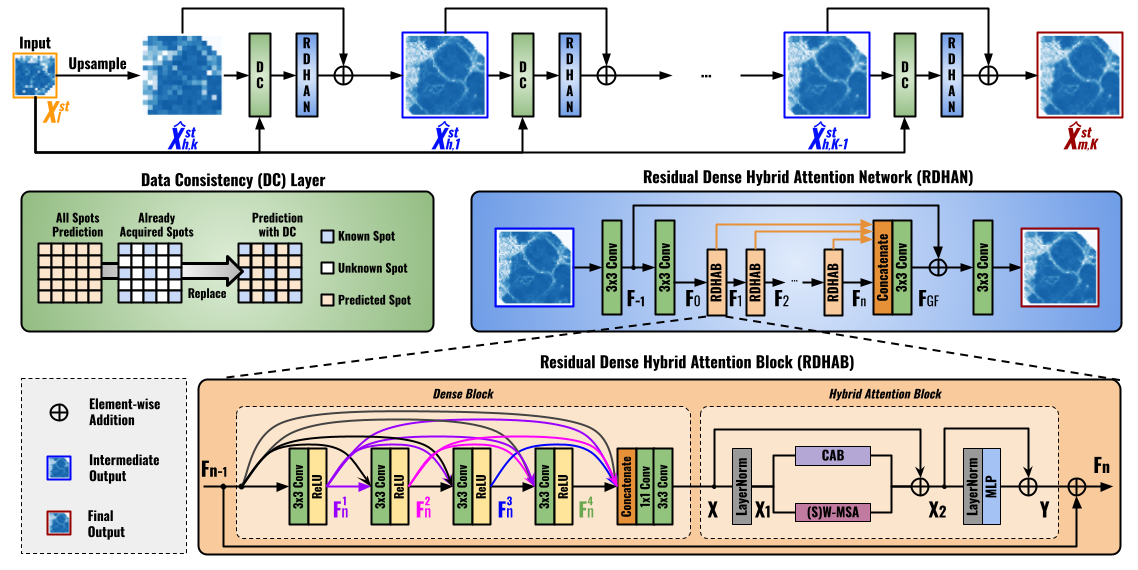}
    \caption{Detailed architecture of the CDCIN used in our S2S-ST framework in Fig. \ref{fig:model-arch}. The model adopts a cascaded structure composed of multiple stages, where each stage refines predictions using a Data Consistency Layer (green box) followed by a Residual Dense Hybrid Attention Network (blue box).}
    \label{fig:network}
\end{figure*}

\subsection{Single-shot Sparser-to-sparse Self-supervised Learning}\label{subsec:image-co-train}
Our single-shot sparser-to-sparse self-supervised learning framework aims to learn a robust ST imputation network (denoted as $f_{\theta}(\cdot)$ to be elaborated in Section \ref{subsec:RDHAN}) with single sparsely sampled ST data and commonly available large-scale natural image datasets. The overall framework is illustrated in Fig. \ref{fig:model-arch}. For the single sparsely sampled ST data, we devise a sparser-to-sparse self-learning strategy. On the other hand, the large-scale natural images with known high resolution ground truth are used for co-training to facilitate the ST imputation learning.

Given a whole slide ST data that contains sparsely sampled spots where non-sampled spots are filled with zero values, we first randomly crop high resolution ST patches ($X_{h}^{st}$) of size $P \times P$. Denoting $M_{h}^{st}$ as a binary sparsely sample pattern mask with the same size of $X_{h}^{st}$, where positive means sparsely sampled spots and zero means non-sampled spots, we apply $M_{h}^{st}$ to $X_{h}^{st}$ to keep only the sparsely sampled spot and generate medium resolution ST patches ($X_{m}^{st}$) of size $P/S \times P/S$, namely sparse ST. $S$ is a scaling factor for subsampling. Then, we further downsample the $X_{m}^{st}$ and generate low resolution ST patches ($X_{l}^{st}$), namely sparser ST. During this process, we also construct a downsampling pattern binary mask ($M_{m}^{st}$) for the sparse to sparser process. Inputting $X_{m}^{st}$ and $X_{l}^{st}$ into our imputation network $f_{\theta}(\cdot)$, we have:
\begin{align}
    \{\hat{X}^{st}_{m,1}, \hat{X}^{st}_{m,2}, \dots, \hat{X}^{st}_{m,K}\} = f_{\theta} (X_{l}^{st}, M_{m}^{st}), \\
    \{\hat{X}^{st}_{h,1}, \hat{X}^{st}_{h,2}, \dots, \hat{X}^{st}_{h,K}\} = f_{\theta} (X_{m}^{st}, M_{h}^{st}),
\end{align}
where $K$ denotes the number of cascade in $f_{\theta}(\cdot)$. $\hat{X}^{st}_{m,K}$ and $\hat{X}^{st}_{l,K}$ means the final imputation outputs predicted from $X_{l}^{st}$ and $X_{m}^{st}$, respectively. With this, our ST self-supervised losses can be formulated as:
\begin{align}
\mathcal{L}^{st}_{m} &= \sum^{k=K}_{k=1}\frac{k}{K}\frac{||(\hat{X}_{{m},k}^{st} - X_{m}^{st}) ||_1}{(P/S)^2}, \\
\mathcal{L}^{st}_{h} &= \sum^{k=K}_{k=1}\frac{k}{K}\frac{||(\mathcal{D}(\hat{X}_{{h},k}^{st}, M_{h}^{st}) - X_{m}^{st})||_1}{(P/S)^2},
\end{align}
where $\mathcal{D}(X,M)$ denotes the downsampling/subsampling operation on $X$ per sampling pattern from $M$. The first loss $\mathcal{L}^{st}_{m}$ is our sparser-to-sparse self-supervised loss, while the second loss $\mathcal{L}^{st}_{h}$ is our sparse-to-dense self-supervised loss. The total loss in the ST domain can thus be formulated as:
\begin{align}
\mathcal{L}^{st} &= \mathcal{L}^{st}_m + \mathcal{L}^{st}_h
\end{align}

For grayscale natural images (GNI) with all high resolution pixel values known, we first randomly crop high resolution patches ($X_{h}^{gni}$) of size \(P \times P\) from the full size images. Similar to the ST data pipeline, based on the same subsampling pattern mask $M_{h}^{st}$ in ST, we apply it to $X_{h}^{gni}$ and generate medium resolution patches ($X_{m}^{gni}$). Then, $X_{m}^{gni}$ is further subsampled per sampling pattern mask $M_{m}^{st}$ to generate low resolution patches ($X_{l}^{gni}$). Inputting $X_{l}^{gni}$ and $X_{m}^{gni}$ into the same imputation network $f_{\theta}(\cdot)$, we have:
\begin{align}
    \{\hat{X}^{gni}_{m,1}, \hat{X}^{gni}_{m,2}, \dots, \hat{X}^{gni}_{m,K}\} = f_{\theta} (X_{l}^{gni}, M_{m}^{st}), \\
    \{\hat{X}^{gni}_{h,1}, \hat{X}^{gni}_{h,2}, \dots, \hat{X}^{gni}_{h,K}\} = f_{\theta} (X_{m}^{gni}, M_{h}^{st}),
\end{align}
and similar here, $\hat{X}^{gni}_{m,K}$ and $\hat{X}^{gni}_{l,K}$ means the final imputation outputs predicted from $X_{l}^{gni}$ and $X_{m}^{gni}$, respectively. With this, our GNI-based fully-supervised co-training losses can be formulated as:
\begin{align}
\mathcal{L}^{gni}_{{m}} &= \sum^{k=K}_{k=1}\frac{k}{K}\frac{||(\hat{X}_{{m},k}^{gni} - X_{m}^{gni})||_1}{(P/S)^2}, \\
\mathcal{L}^{gni}_{h} &= \sum^{k=K}_{k=1}\frac{k}{K}\frac{||(\hat{X}_{{h},k}^{gni} - X_{h}^{gni})||_1}{P^2},
\end{align}
where the first loss $\mathcal{L}^{gni}_{m}$ is our sparser-to-sparse co-training loss, while the second loss $\mathcal{L}^{gni}_{h}$ is our sparse-to-dense co-training loss. Thus, the total loos in the GNI domain can be formulated as:
\begin{align}
\mathcal{L}^{gni} &= \mathcal{L}^{gni}_{m} + \mathcal{L}^{gni}_{h}
\end{align}

Finally, the total loss \(\mathcal{L}\) is then defined as a weighted combination of ST loss and GNI loss:
\begin{align*}
\mathcal{L} &= \lambda \cdot \mathcal{L}^{st} + \mathcal{L}^{gni},
\end{align*}
where $\lambda$ is weighting coefficients that balance the contributions of the ST gene expression data and GNI data, respectively. The grayscale natural image loss $\mathcal{L}^{gni}$ ensures the model's fidelity to the general image data imputation, while the gene loss $\mathcal{L}_{st}$ enforces alignment with the spatial transcriptomics data. 
Notably, the loss computation for $L^{st}$ excludes spots located outside the tissue region.

During inference, the trained $f_{\theta}(\cdot)$ is used to predict a whole slide high resolution ST profile from sparsely sampled ST data in a sliding window manner (Fig. \ref{fig:model-arch}, bottom left). The final whole slide output is reconstructed by stitching predicted patches into their original positions. For overlapping regions, the final value is computed as the weighted average of all predictions covering that region.

\subsection{Cascaded Data Consistent Imputation Network} \label{subsec:cascade}
Our Cascaded Data Consistent Imputation Network (CDCIN) is designed to keep sampled sparse data during imputation, especially for sparse ST gene expression. It integrates a cascaded data consistency framework to maintain the integrity and coherence of imputation across the network. The network consists of $K$ stages, each of that performs a Data Consistency (DC) operation followed by a Residual Dense Hybrid Attention Network (RDHAN). Before being fed into the model, the relative low resolution (LR, i.e. $X_{m}^{st}$) input is upsampled to match the relative high resolution (HR, i.e. $X_{h}^{st}$) size. 

The DC module continuously reintroduces the known data during the intermediate processes of the model to guide the model. The DC operation acts as a regularization step, ensuring that the imputation results remain consistent and preventing information degradation during the cascading process. For example, the DC operation can be expressed as:
\begin{equation}
\text{DC}(X^{st}_{h},M^{st}_{h}) =  \hat{X}^{st}_{h,k} \odot (1 - M^{st}_{h,DC}) + X^{st}_{h} \odot M^{st}_{h},
\end{equation}
where \(M^{st}_{h}\) is a binary mask indicating the positions of LR (i.e. $X^{st}_{m}$) inputs corresponding to their locations in the HR (i.e. $X^{st}_{h}$) data. 

The data processed by the DC module is then fed into the RDHAN for data imputation and completion. Through its cascaded design, the network operates in a step-by-step manner, iteratively refining and improving the imputation results at each stage, ensuring progressively enhanced accuracy and coherence. Mathematically, the cascade structure can be represented as:
\begin{align}
\hat{X}^{st}_{h,0} &= \mathcal{U}({X^{st}_{m}}), \\
\hat{X}^{st}_{h,k} &= \hat{X}^{st}_{h,k-1} + \text{RDHAN}(\text{DC}(\hat{X}^{st}_{h,k-1}, M_{h}^{st})), \nonumber \\
&\phantom{=} \hspace{3cm}  \text{for } k = 1, 2, \dots, K,
\end{align}

where $\mathcal{U}(X)$ denotes the upsampling operation, explicitly resizing the input data from a lower resolution shape to the corresponding higher resolution shape.

\subsection{Residual Dense Hybrid Attention Network}\label{subsec:RDHAN}
Our proposed Residual Dense Hybrid Attention Network (RDHAN), as illustrated in Fig.~\ref{fig:network}, is built upon a cascade of Residual Dense Hybrid Attention Block (RDHAB) units, which are designed to effectively capture both local and global feature dependencies for enhanced reconstruction performance. At the final stage, the outputs of all RDHAB modules are concatenated to form a comprehensive global feature representation, enabling the model to reconstruct high quality outputs with rich spatial and contextual information.

Each Residual Dense Hybrid Attention Block (RDHAB) is composed of a Residual Dense Block (RDB) and a Hybrid Attention Block (HAB). The RDB leverages dense connections to facilitate feature reuse and gradient flow, combined with residual learning to stabilize training. This design enables the extraction of rich hierarchical features from local regions, which are crucial for capturing fine-grained details in complex tasks. The HAB is designed to combine the strengths of channel attention and Swin Transformer-based spatial attention. Specifically, we introduce a Channel Attention-based Convolution Block (CAB) into the standard Swin Transformer block. As illustrated in Fig.~\ref{fig:network}, the CAB is inserted after the first LayerNorm (LN) layer and operates in parallel with the Window-based Multi-head Self-Attention (W-MSA) module. To mitigate potential optimization conflicts between the CAB and W-MSA, we scale the output of the CAB by a small constant \(\alpha\) \citep{chenActivatingMorePixels2023}. Given an input feature \(X\), the computation within the HAB is formulated as follows:
\begin{align}
X_1 &= \text{LN}(X), \\
X_2 &= \text{(S)W-MSA}(X_1) + \alpha \cdot \text{CAB}(X_1) + X, \\
Y &= \text{MLP}(\text{LN}(X_2)),
\end{align}
where \(X_i\) denotes intermediate features, \(\text{LN}\) represents layer normalization, \(\text{(S)W-MSA}\) refers to the standard and shifted window multi-head self-attention modules, \(\text{MLP}\) is a multi layer perceptron, and \(Y\) is the final output of the HAB.

The channel attention mechanism in the CAB enhances the model's ability to prioritize important feature channels, while the Swin Transformer-based spatial attention captures long range dependencies within the feature maps. By integrating these two mechanisms, the HAB effectively captures both local and global feature dependencies, significantly improving the model's representation and reconstruction capabilities. This makes the HAB particularly suitable for complex tasks such as image inpainting and high resolution reconstruction.

\subsection{Datasets}\label{subsec:datasets}
This study employs a validation strategy using ST data spanning diverse tissue types and pathological conditions, enabling rigorous evaluation across multiple contexts. Our dataset includes five Xenium breast cancer samples comprising two invasive lobular carcinomas (ILC: TENX94, TENX96) and three invasive ductal carcinomas (IDC: TENX95, TENX97, TENX98). The collection further contains prostate cancer samples (TENX157), healthy liver tissue (TENX121), and diseased lymphoid tissue (TENX143). All datasets curated from HEST-1K \citep{jaumeHEST1kDatasetSpatial2024}. For natural image co-training, we incorporated DIV2K \citep{agustssonNTIRE2017Challenge2017}, the benchmark super resolution dataset containing 800 training and 100 validation image pairs at 2K resolution.

The preprocessing pipeline consists of two parallel branches, which handle ST gene expression data and grayscale natural image data, respectively. For each gene channel, we normalized the expression value via log-normalization, as standard value pre-processing as done in previous works \citep{zhaoDISTSpatialTranscriptomics2023a}. Then, the data is cropped into high resolution (\(X_{h}^{st}\)) patches of size \(P \times P\) with a stride of \(S\), augmented through random horizontal and vertical flipping and random rotation by multiples of 90 degrees (0$^\circ$, 90$^\circ$, 180$^\circ$, and 270$^\circ$) during training. The high resolution ST data (\(X_{h}^{st}\)) is then downsampled with stride \(S\) to generate medium resolution patches \(X_{m}^{st}\) (size \(P/S \times P/S\)) and low resolution patches \(X_{l}^{st}\) (size \(P/S^2 \times P/S^2\)). Binary masks \(M_{h}^{st}\) and \(M_{m}^{st}\) are generated to indicate selected spots after downsampling, which are also applied to the GNI branch.
For grayscale natural image preprocessing, RGB images are converted to grayscale and normalized to the range [0, 1]. Data augmentation is performed via random horizontal and vertical flipping, and random rotations by arbitrary angles within the range [0$^\circ$, 360$^\circ$). High resolution patches \(X_{h}^{gni}\) of size \(P \times P\) are then cropped from the images and subsequently downsampled to generate medium resolution patches \(X_{m}^{gni}\) (size \(P/S \times P/S\)) and low resolution patches \(X_{l}^{gni}\) (size \(P/S^2 \times P/S^2\)), consistent with the ST branch preprocessing steps.

\subsection{Implementation Details}\label{subsec:implement}
ST gene expression data is cropped into patches of size $P \times P$ with a stride of $S$, where $P = 64$ and $S = 2$, while natural images are directly cropped into $P \times P$ patches without stride. Downsampling is performed by selecting the top-left pixel of each $S \times S$ grid, reducing the spatial resolution by a factor of $S$. 

Considering the balance among computational speed, memory consumption, and model performance (extensively evaluated in Section \ref{sec:results}), by default, our model employs a cascaded architecture of 3 DC-RDHAN networks in series, each consisting of 8 RDHAB blocks. The RDB is configured with 32 initial channels, a dense connection growth rate of 32, and 4 convolutional layers per block. In the HAB, the window-based self-attention mechanism uses a window size of 8, and the CAB output weight $\alpha$ is set to 0.01.

The gene loss weight $\lambda$ is set to 10, and the initial learning rate is $1 \times 10^{-4}$. The model is trained for 3000 epochs using the Adam optimizer with default parameters on a single NVIDIA A6000 GPU. Training is performed on one gene at a time, focusing on specific gene targets.

\subsection{Baselines and Evaluation Metrics}\label{subsec:baseline+metrics}
For a fair comparison, we selected two methods that also utilize only ST gene expression data as input for enhancing gene expression resolution: BayesSpace \citep{zhaoSpatialTranscriptomicsSubspot2021} and DIST \citep{zhaoDISTSpatialTranscriptomics2023a}. Additionally, we included TESLA \citep{huDecipheringTumorEcosystems2023} as an additional baseline method, which further integrates histology images along with ST gene expression to improve spatial gene expression resolution.

To evaluate the performance of our proposed model, we employ three widely used metrics: Mean Absolute Error (MAE), Pearson Correlation Coefficient (PCC), and Structural Similarity Index (SSIM). These metrics are computed on gene expression data to provide a comprehensive assessment of the model's performance.

\section{Experimental Results}\label{sec:results}
\begin{table}[tbh!]
\centering
\footnotesize
\caption{Quantitative comparison of different models on eight spatial transcriptomics datasets across various tissue types and genes. The best performance for each dataset is highlighted in bold.}\label{tab:model-compare}
  \resizebox{\linewidth}{!}{
    \begin{tabular}{c c c c c c c c}
    \toprule
    \textbf{Dataset} & \textbf{Gene} & \textbf{Model} & \textbf{MAE ($\downarrow$)} & \textbf{PCC ($\uparrow$)} & \textbf{SSIM ($\uparrow$)} \\
    
    \midrule
    \multirow{4}{*}{\shortstack{TENX94\\Breast\\Cancer}} 
    & \multirow{4}{*}{ERBB2} 
        & TESLA                     & 0.5702            & 0.7704            & 0.5581 \\
     &  & BayesSpace                & 0.5925            & 0.7802            & 0.6187 \\
     &  & DIST                      & 0.4470            & 0.8467            & 0.7107 \\
     &  & \textbf{CDCIN (Ours)}     & \textbf{0.3840}   & \textbf{0.8539}   & \textbf{0.7613} \\

    \midrule
    \multirow{4}{*}{\shortstack{TENX95\\Breast\\Cancer}} 
    & \multirow{4}{*}{ERBB2} 
        & TESLA                     & 0.4035            & 0.6186            & 0.6719 \\
     &  & BayesSpace                & 0.3922            & 0.7047            & 0.7021 \\
     &  & DIST                      & 0.3741            & 0.7103            & 0.7067 \\
     &  & \textbf{CDCIN (Ours)}     & \textbf{0.2801}   & \textbf{0.8025}   & \textbf{0.7764} \\
    
    \midrule
    \multirow{4}{*}{\shortstack{TENX96\\Breast\\Cancer}} 
    & \multirow{4}{*}{ERBB2} 
        & TESLA                     & 0.5676            & 0.7805            & 0.5425 \\
     &  & BayesSpace                & 0.5704            & 0.7843            & 0.6044 \\
     &  & DIST                      & 0.4430            & 0.8506            & 0.6858 \\
     &  & \textbf{CDCIN (Ours)}     & \textbf{0.3872}   & \textbf{0.8549}   & \textbf{0.7476} \\
    
    \midrule
    \multirow{4}{*}{\shortstack{TENX97\\Breast\\Cancer}} 
    & \multirow{4}{*}{ERBB2} 
        & TESLA                     & 0.4376            & 0.6197            & 0.6453 \\
     &  & BayesSpace                & 0.4269            & 0.6933            & 0.6748 \\
     &  & DIST                      & 0.3629            & 0.7563            & 0.7231 \\
     &  & \textbf{CDCIN (Ours)}     & \textbf{0.2965}   & \textbf{0.8164}   & \textbf{0.7691} \\
    
    \midrule
    \multirow{4}{*}{\shortstack{TENX98\\Breast\\Cancer}} 
    & \multirow{4}{*}{ERBB2} 
        & TESLA                     & 0.6058            & 0.6495            & 0.4567 \\
     &  & BayesSpace                & 0.5517            & 0.7309            & 0.5591 \\
     &  & DIST                      & 0.4452            & \textbf{0.8085}   & 0.6517 \\
     &  & \textbf{CDCIN (Ours)}     & \textbf{0.4081}   & 0.7992            & \textbf{0.6802} \\
    
    \midrule
    \multirow{4}{*}{\shortstack{TENX121\\Liver\\Healthy}} 
    & \multirow{4}{*}{CYP2A7} 
        & TESLA                     & 0.2969            & 0.5066            & 0.6841 \\
     &  & BayesSpace                & 0.2691            & 0.5800            & 0.7496 \\
     &  & DIST                      & 0.2367            & 0.5956            & 0.7622 \\
     &  & \textbf{CDCIN (Ours)}     & \textbf{0.1889}   & \textbf{0.7243}   & \textbf{0.8247} \\
    
    \midrule
    \multirow{4}{*}{\shortstack{TENX143\\Lymphoid\\Diseased}} 
    & \multirow{4}{*}{XBP1} 
        & TESLA                     & 0.6896            & 0.6090            & 0.6061 \\
     &  & BayesSpace                & 0.7725            & 0.5529            & 0.5723 \\
     &  & DIST                      & 0.6546            & 0.6422            & 0.6206 \\
     &  & \textbf{CDCIN (Ours)}     & \textbf{0.5578}   & \textbf{0.6865}   & \textbf{0.7120} \\
    
    \midrule
    \multirow{4}{*}{\shortstack{TENX157\\Prostate\\Cancer}} 
    & \multirow{4}{*}{CCN1} 
        & TESLA                     & 0.7179            & 0.3624            & 0.7571 \\
     &  & BayesSpace                & 0.7616            & 0.4075            & 0.7925 \\
     &  & DIST                      & 0.6661            & 0.4700            & 0.8174 \\
     &  & \textbf{CDCIN (Ours)}     & \textbf{0.5264}   & \textbf{0.6224}   & \textbf{0.8571} \\
     
    \bottomrule
    \end{tabular}
}
\end{table}

\begin{figure*}[htb!] 
    \centering
    \includegraphics[width=0.60\textwidth]{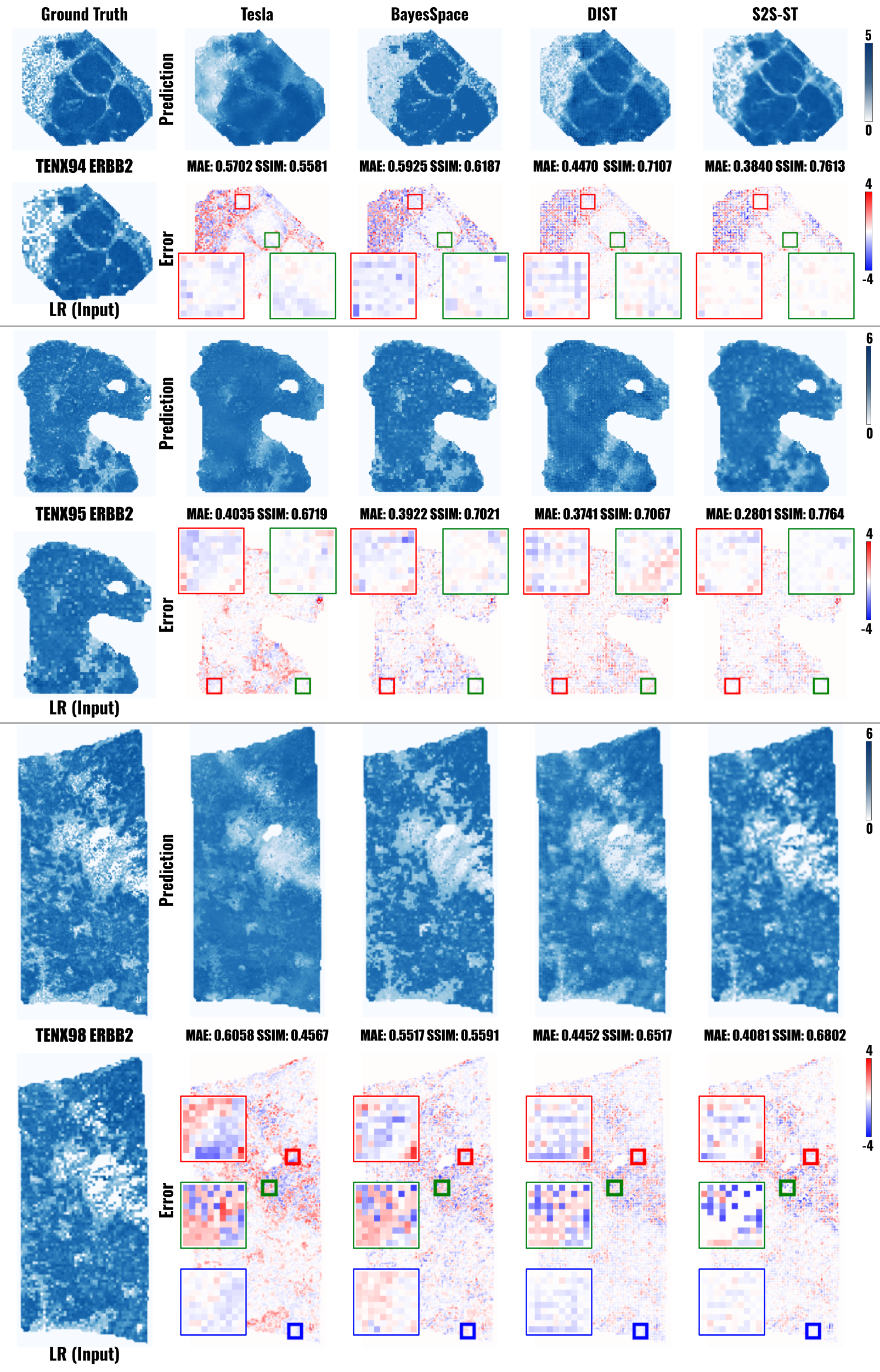}
    \caption{Qualitative comparison of spatial gene expression reconstruction for the ERBB2 gene across three breast cancer samples (TENX94, TENX95, and TENX98). Each row shows results from five competing methods: Tesla, BayesSpace, DIST, and the proposed CDCIN, evaluated against the ground truth and low resolution (LR) input. For each method, predicted expression maps are shown alongside error maps (difference from ground truth), with inset regions highlighting local reconstruction accuracy. Quantitative metrics (MAE and SSIM) are provided under each prediction. The CDCIN model consistently yields lower reconstruction error and clearer structural detail across all samples, demonstrating its superior capability for accurate spatial imputation.}
    \label{fig:model-compare}
\end{figure*}

\begin{figure*}[htb!] 
    \centering    
    \includegraphics[width=0.7\textwidth]{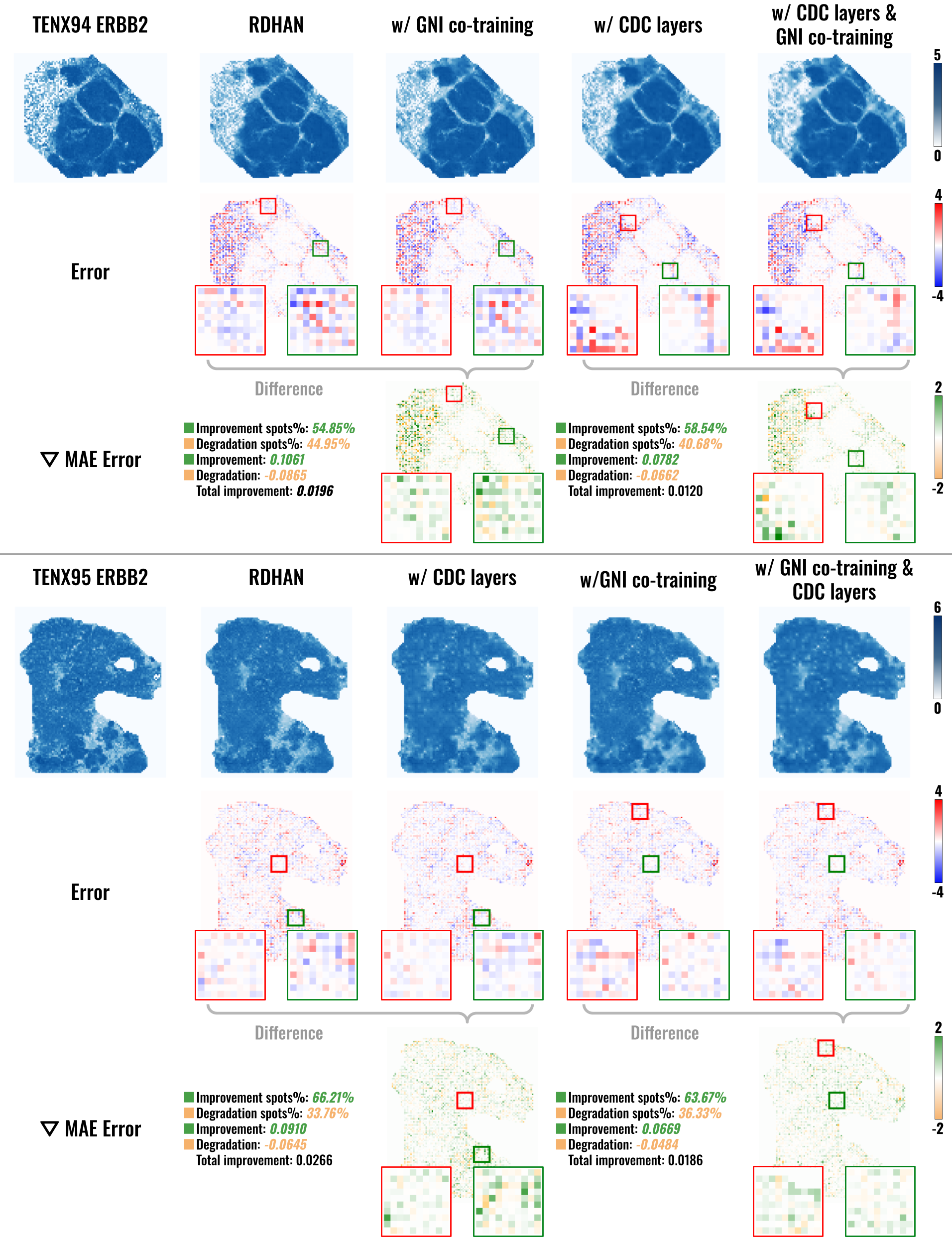}
    \caption{Ablation study evaluating the impact of key components in the proposed CDCIN model using ERBB2 expression in breast cancer samples TENX94 and TENX95. Each row compares the reconstruction performance of the base RDHAN model with variants incorporating GNI co-training, CDC layers, or both. Top panels show the reconstructed expression maps, middle panels show absolute error maps relative to the ground truth, and bottom panels display MAE error difference maps, highlighting regions of improvement or degradation. Quantitative indicators (improvement ratio, average gain, and total improvement) are provided for each variant. Results demonstrate that both GNI co-training and CDC layers contribute positively to performance, with the combination yielding the most consistent error reduction and structural fidelity.}
    \label{fig:model_ablation}
\end{figure*}

\begin{figure*}[htb!] 
    \centering
    \includegraphics[width=0.95\textwidth]{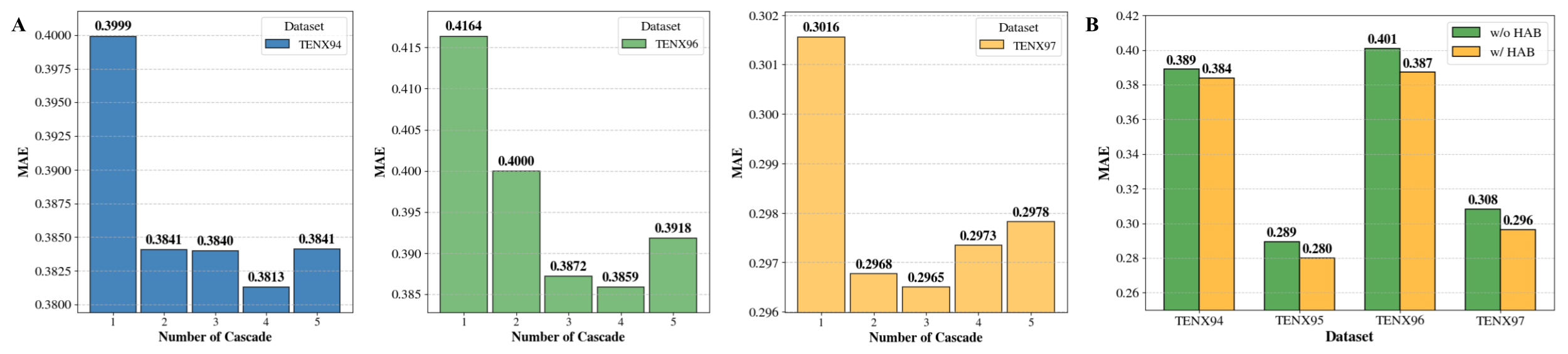}
    \caption{Ablation study on the effect of cascade depth and the Hybrid Attention Block (HAB) in the CDCIN model. 
    (A) MAE comparison across different numbers of cascaded stages (from 1 to 5) on datasets TENX94, TENX96, and TENX97. Performance improves significantly up to three cascades, after which gains saturate or slightly decline, indicating that three cascades provide a reasonable trade-off between accuracy and computational cost. 
    (B) MAE comparison of models with and without the Hybrid Attention Block (HAB) on four datasets. Incorporating HAB consistently reduces MAE, demonstrating its contribution to feature refinement and spatial accuracy.}
    \label{fig:cascade_ablation}
\end{figure*}

To rigorously evaluate our proposed CDCIN model, we conducted comprehensive comparisons across multiple datasets with current state-of-the-art (SOTA) methods, including Tesla, BayesSpace, and DIST, in enhancing the resolution of spatial gene expression. As illustrated in Fig.\ref{fig:model-compare}, our CDCIN model can consistently surpass these existing approaches, delivering the lowest MAE and highest SSIM across different ST samples. Taking TENX94 ERBB2 as a qualitative comparison example, the original LR inputs exhibit significant noise and indistinct structural details due to reduced spot sampling. In comparison, the Tesla model tends toward excessive smoothing and blurring, significantly diminishing crucial details (e.g., MAE of 0.5702 and SSIM of 0.5581). BayesSpace provides minimal effective improvement in resolution, with relatively higher errors and moderate SSIM scores (e.g., MAE of 0.5925 and SSIM of 0.6187). The DIST model improves accuracy but still struggles in areas of rapid data variation, showing noticeable local discrepancies (e.g., MAE of 0.4470 and SSIM of 0.7107). On the other hand, our CDCIN model adeptly preserves intricate structural details, achieving notably superior performance (MAE of 0.3840 and SSIM of 0.7613). Similar observations can be consistently found in the TENX95 and TENX98 datasets, which underscore CDCIN's superior capability in finely enhancing resolution and effectively reducing noise.

Furthermore, as shown in Table \ref{tab:ablation_consolidated}, the proposed CDCIN model demonstrates robust and consistent performance across a broader spectrum of tissue types and conditions, including healthy (e.g., TENX121 liver), diseased (TENX143 lymphoid), and various cancerous tissues (TENX157 prostate cancer). Specifically, in the TENX121 healthy liver dataset, CDCIN significantly outperforms other methods with markedly lower MAE (0.1889) and higher SSIM (0.8247), highlighting its effectiveness in normal tissue contexts. Likewise, for the TENX143 lymphoid disease dataset, CDCIN exhibits superior performance, obtaining the lowest MAE (0.5578) and highest SSIM (0.7120), reflecting its strong predictive capacity in diseased states. Similarly, CDCIN achieves the best results in the prostate cancer dataset TENX157, with MAE (0.5264) and SSIM (0.8571), indicating its capability to accurately capture gene expression variations in cancerous tissues. These diverse results further emphasize CDCIN’s exceptional versatility, robustness, and generalizability across different health states and tissue types.

\begin{table*}[tbh]
\centering
\footnotesize
\caption{Comprehensive ablation study results evaluating the impact of GNI co-training and Cascade Data Consistency (CDC).}
\label{tab:ablation_consolidated}
\begin{tabular}{c c c c c c}
\toprule
\textbf{Dataset} & \textbf{Variant} & \textbf{MAE} & \textbf{PCC} & \textbf{SSIM} & \textbf{Improvement} \\
\midrule
\multirow{4}{*}{\shortstack{TENX94\\Breast\\Cancer}} 
& RDHAN (w/o both) & 0.4226 & 0.8216 & 0.7212 & Baseline \\
& w/ GNI co-training & 0.4030 & 0.8308 & 0.7380 & 4.6\% $\downarrow$MAE, 1.1\% $\uparrow$PCC, 2.3\% $\uparrow$SSIM \\
& w/ CDC & 0.3959 & 0.8426 & 0.7508 & 6.3\% $\downarrow$MAE, 2.6\% $\uparrow$PCC, 4.1\% $\uparrow$SSIM \\
& CDCIN (w/ both) & 0.3840 & 0.8539 & 0.7613 & 9.1\% $\downarrow$MAE, 3.9\% $\uparrow$PCC, 5.6\% $\uparrow$SSIM \\
\midrule

\multirow{4}{*}{\shortstack{TENX95\\Breast\\Cancer}}
& RDHAN (w/o both) & 0.3164 & 0.7581 & 0.7352 & Baseline \\
& w/ GNI co-training & 0.2987 & 0.7906 & 0.7568 & 5.6\%\,\,\, $\downarrow$MAE, 4.3\% $\uparrow$PCC, 2.9\% $\uparrow$SSIM \\
& w/ CDC & 0.2898 & 0.7933 & 0.7721 & 8.4\%\,\,\, $\downarrow$MAE, 4.6\% $\uparrow$PCC, 5.0\% $\uparrow$SSIM \\
& CDCIN (w/ both) & 0.2801 & 0.8025 & 0.7764 & 11.5\% $\downarrow$MAE, 5.9\% $\uparrow$PCC, 5.6\% $\uparrow$SSIM \\
\midrule

\multirow{4}{*}{\shortstack{TENX96\\Breast\\Cancer}}
& RDHAN (w/o both) & 0.4205 & 0.8296 & 0.7154 & Baseline \\
& w/ GNI co-training & 0.4031 & 0.8449 & 0.7339 & 4.1\% $\downarrow$MAE, 1.8\% $\uparrow$PCC, 2.6\% $\uparrow$SSIM \\
& w/ CDC & 0.4039 & 0.8434 & 0.7294 & 4.0\% $\downarrow$MAE, 1.7\% $\uparrow$PCC, 2.0\% $\uparrow$SSIM \\
& CDCIN (w/ both) & 0.3872 & 0.8549 & 0.7476 & 7.9\% $\downarrow$MAE, 3.1\% $\uparrow$PCC, 4.5\% $\uparrow$SSIM \\
\midrule

\multirow{4}{*}{\shortstack{TENX97\\Breast\\Cancer}}
& RDHAN (w/o both) & 0.3380 & 0.7637 & 0.7201 & Baseline \\
& w/ GNI co-training & 0.3150 & 0.7952 & 0.7508 & 6.8\%\,\,\, $\downarrow$MAE, 4.1\% $\uparrow$PCC, 4.3\% $\uparrow$SSIM \\
& w/ CDC & 0.3013 & 0.8043 & 0.7629 & 10.9\% $\downarrow$MAE, 5.3\% $\uparrow$PCC, 5.9\% $\uparrow$SSIM \\
& CDCIN (w/ both) & 0.2965 & 0.8164 & 0.7691 & 12.3\% $\downarrow$MAE, 6.9\% $\uparrow$PCC, 6.8\% $\uparrow$SSIM \\
\bottomrule
\end{tabular}
\end{table*}

We also conducted a range of ablative studies to investigate the contribution of different components of CDCIN, as follows.

\noindent\textbf{Impact of GNI Co-learning:}
Ablation studies underscore the systematic benefits introduced by GNI co-training, as shown through detailed evaluations on the TENX94 dataset (Fig.~\ref{fig:model_ablation}). The model learns resolution enhancing patterns that significantly improve ST gene expression prediction accuracy, particularly in regions previously associated with high error rates. Statistical analysis indicates that 54.85\% of spatial locations exhibit improved accuracy, with a mean error reduction of 0.1061, which clearly outweighs the marginal degradation (0.0865) observed in the remaining regions. This corresponds to a net MAE improvement of 0.0196. While GNI co-training still leads to improvements when combined with Cascade Data Consistency (CDC) layers, the additional MAE gain on TENX94 is relatively smaller (0.0120), as the CDC layers have already contributed significantly to the performance improvement. Moreover, the advantages of GNI co-training generalize consistently across all evaluated datasets (Table~\ref{tab:ablation_consolidated}), where it achieves a reliable MAE reduction of 4–7\%. These results highlight the robustness and general utility of incorporating GNI-based co-learning for spatial gene expression enhancement.

\noindent\textbf{Impact of Cascade Data Consistency:}
Complementing the improvements brought by GNI, CDC layers emerge as equally vital components. Among all tested configurations, CDC alone delivers the most considerable performance gains, as exemplified by results on TENX95 (Fig.~\ref{fig:model_ablation}). These layers simultaneously mitigate local prediction errors while faithfully preserving informative patterns present in the input, ensuring that refined outputs remain anchored in biologically plausible signals. Quantitatively, 66.27\% of spots demonstrate improved prediction accuracy, with a net MAE gain of 0.0266. Visual analyses of error maps further reveal that CDC layers effectively smooth out local inconsistencies while retaining spot level integrity, thereby achieving a dual objective: enhancing precision while maintaining structural consistency. Although the integration of GNI co-training with CDC layers introduces minor interference (reflected by a slightly lower combined MAE gain of 0.0186 compared to the 0.0266 achieved by CDC alone on TENX95), their joint application ultimately yields synergistic improvements. Finally, dataset wide evaluations underscore the broader effectiveness of CDC, with results showing up to 10.9\% MAE reduction, 5.3\% PCC improvement, and 5.9\% SSIM enhancement over baseline (Table~\ref{tab:ablation_consolidated}). These findings reinforce the critical role of CDC, especially when integrated with GNI co-learning strategies.

\noindent\textbf{Impact of Number of Cascades}
As illustrated in Fig.~\ref{fig:cascade_ablation}, the number of Cascade Data Consistency (CDC) modules under GNI co-training conditions plays a pivotal role in determining model performance. Our experiments reveal a positive correlation between cascade depth and performance metrics up to three CDC layers. Specifically, the model achieves optimal predictive accuracy with three CDC layers, beyond which further increases in depth yield diminishing returns. Balancing performance gains with computational cost, we select the three layer configuration as the optimal setting, offering an effective trade-off between complexity and accuracy.

\noindent\textbf{Impact of HAB}
As shown in Fig.~\ref{fig:cascade_ablation}, the incorporation of HAB leads to consistent performance improvements across all datasets, as evidenced by the reduction in MAE. While the degree of improvement varies—with more notable gains observed on TENX95 and TENX97—the overall trend highlights HAB’s effectiveness in enhancing the model’s representational capacity. The relatively smaller improvements on TENX94 and TENX96 suggest a saturation point where the model already captures considerable structural information. Nonetheless, the consistent downward shift in MAE demonstrates that HAB contributes positively and robustly to predictive accuracy across diverse data scenarios.

\section{Discussion}
Our proposed Single-shot S2S-ST model introduces a data-efficient framework for ST gene expression imputation by jointly training on a single ST sample and a large corpus of natural images—a strategy that, to the best of our knowledge, has not been previously explored. Through cross-modal co-learning, the model achieves robust sample-specific adaptation with minimal supervision, setting a new state of the art. This breakthrough directly confronts two persistent challenges in the field: (i) the exorbitant costs associated with high resolution ST data generation, and (ii) the chronic insufficiency of publicly available datasets—both in terms of quantity and biological diversity. By pioneering the integration of naturally abundant image data as a complementary knowledge source, our approach effectively overcomes the data scarcity bottleneck that has long hampered ST computational methodologies.

This advancement is grounded in three key innovations that collectively enable our model’s data-efficient yet high fidelity performance. First, we introduce sparser-to-sparse learning, a self-supervised framework that departs from traditional data-intensive paradigms. By respecting the intrinsic spatial geometry of biological systems, it enables learning directly from sparse measurements without requiring dense supervision. Second, we leverage cross-domain co-learning with natural images, uncovering deep structural analogies between gene expression patterns and natural textures. The observed transferability, reflecting structural parallels between ST gene expression imputation and image modeling, results in a significant MAE reduction (Table~\ref{tab:ablation_consolidated}) and highlights their effectiveness in cross-domain learning. Third, our cascaded data-consistent architecture enforces iterative refinement with memory, progressively enhancing resolution while strictly preserving original measurements at every stage. This mechanism ensures biologically grounded reconstructions, where known data serve as anchors and context-aware imputation fills in the gaps.

Despite its advantages, our method has limitations that open avenues for future work. We discuss these aspects below, focusing on computational constraints, model scope, and potential extensions.  

First, the necessity for sample-specific training constitutes a primary constraint of our framework. While achieving superior reconstruction fidelity, each new dataset requires complete model retraining, with the current implementation typically demanding six hours of GPU computation. This duration, though reasonable for research settings, may hinder clinical translation where rapid turnaround is essential. Future improvements could explore (1) distilled network architectures maintaining accuracy while reducing parameters\citep{moslemiSurveyKnowledgeDistillation2024}, or (2) meta-learning approaches enabling knowledge transfer across similar tissue specimens\citep{gevaertMetalearningReducesAmount2021}. The current paradigm deliberately favors reconstruction precision over operational expediency—a strategic compromise that future hardware advances may help reconcile.

Second, our methodology currently operates on a selective gene panel rather than full transcriptome coverage, which while computationally efficient may overlook biologically relevant interactions. This design choice was motivated by the trade-off between computational tractability and biological completeness, particularly when working with high resolution spatial data. Future iterations could adopt a multi-channel architecture that simultaneously processes distinct gene groups, thereby achieving comprehensive coverage without proportionally increasing training time. Such an extension would need to address the inherent challenge of maintaining inter-gene correlation structures while scaling across thousands of genomic features. The present focused approach, nevertheless, provides a robust foundation for these potential expansions, having established effective frameworks for spatial pattern preservation.

Third, our evaluation focused exclusively on technical metrics (PCC/MAE/SSIM) for expression reconstruction, without validating performance in downstream biological analyses. While these measures rigorously quantify spatial pattern preservation and expression value accuracy, effects on specific downstream applications such as differential expression testing or cell-type mapping are still needed in the future. Future work will also include systematic benchmarking using established bioinformatics workflows to assess whether the observed improvements in technical metrics translate to tangible benefits in biological discovery. 

Lastly, while our current framework focuses on 2D spatial analysis, emerging 3D spatial transcriptomics technologies present both new challenges and opportunities\citep{almagro-perezAIdriven3DSpatial2025}. The methodology could be naturally extended to volumetric data by redefining the sparse sampling paradigm along the z-axis, where central slices might be fully sequenced while adjacent layers use sparse measurement. This adaptation would require addressing several key considerations: computational scalability for high-volume 3D datasets, development of volumetric attention mechanisms to capture spatial relationships in all dimensions, and novel normalization approaches to account for depth-dependent signal attenuation. Particularly promising would be an asymmetric sampling strategy where the xy-plane maintains high resolution while applying strategic sparsity along the z-axis. Such a 3D implementation could prove transformative for studying thick tissue sections and organoids while potentially reducing total sequencing costs through optimized sparse sampling in all three dimensions.

\section{Conclusion}

This paper presents Single-shot S2S-ST, an innovative framework designed to address the data scarcity challenge in spatial transcriptomics analysis without compromising reconstruction accuracy. Through the integration of cross-modal learning with a novel sparser-to-sparse training paradigm, Single-shot S2S-ST achieves a balance between data efficiency and performance by effectively leveraging natural image knowledge. The cascaded reconstruction architecture with data-consistent refinement further enhances the reliability of predictions, demonstrating significant improvement in imputation accuracy compared to existing methods. The comprehensive evaluation of Single-shot S2S-ST on multiple tissue types highlights its ability to achieve robust performance across various biological contexts while requiring minimal training samples.




\bibliographystyle{model2-names.bst}\biboptions{authoryear}
\bibliography{refs}

\end{document}